\documentclass{llncs}

\usepackage[x]{optional}

\newcommand{\subject}{CSLP 2004}

\newcommand{\keywords}{Constraint Solving and Language Processing}

\newcommand{\jv}{jv@ruc.dk}

\usepackage{amssymb,amsmath}

\usepackage[pagebackref,bookmarks=false]{hyperref}

\makeatletter
\newcommand\o@maketitle{} \let\o@maketitle\maketitle \def\maketitle{\hypersetup{pdftitle={\@title},pdfauthor={\@author},pdfsubject={\subject},pdfkeywords={\keywords}}\o@maketitle}
\makeatother

\usepackage{latexsym}

\let\M=\mbox

\def\eurotoday{\number\year--\ifnum\month<10 0\fi
\number\month--\ifnum\day<10 0\fi\number\day}

\newdimen\lineindent \lineindent=1cm

\def\cB{{\cal B}}

\def\cS{{\cal S}}
\def\cT{{\cal T}}

\def\cV{{\cal V}}

\def\C#1{\M{$\bf #1$}}

\def\T#1{\M{\sf #1}}

\def\CONCAT{\M{\^{}}}

\def\E{\M{$\epsilon$}}

\def\TAB{\makebox[\lineindent]{}}

\def\TABX{\makebox[.5\lineindent]{}}

\def\LEAD{\makebox[4ex][l]{$\leadsto$}}

\def\LINEX#1#2{\TABX\hspace{#1\lineindent}$#2$ \hfill \TABX}

\def\LINEY#1#2#3{\TABX \hfill $#1 ~\succ~ #2$ ~~~~~~~ #3 \TABX}

\def\LINE#1#2#3#4{\TABX\hspace{#1\lineindent}$#2 ~\succ~ #3$ \hfill #4 \TABX}

\newcommand{\BS}{\raisebox{.25ex}{\hspace*{0em}\small$\backslash$}}

\newcommand{\FS}{\raisebox{.25ex}{\hspace*{0em}\small$/$}}

\def\CIRC{\M{$\circ$}}
\def\BIGCIRC{\M{$\bigcirc$}}
\def\BULLET{\M{$\bullet$}}

\def\SYN#1{\M{$\lceil #1 \rceil$}}
\def\SEM#1{\M{$\lfloor #1 \rfloor$}}
\def\SYNSEM#1{\M{$| #1 |$}}
\def\INT#1{\M{$\lbrack\!\lbrack #1 \rbrack\!\rbrack$}}

\def\CAT#1#2{\TAB\T{#1}~~:~~$ #2 $}

\def\DEFX#1#2#3{\TAB\C{#1}~~~$\equiv$~~~$ #2 $~~$|$~~$ #3 $}

\def\DEFXX#1#2#3#4{\TAB\C{#1}~~~$\equiv$~~~$ #2 $~~$|$~~$ #3 $~~$|$~~$ #4 $}

\def\TREE{\begin{Tree}}

\def\endTREE{\end{Tree}\begin{center}\usebox{\TeXTree}\end{center}}

\begin{document}

\title{Multi-dimensional Type Theory: \\ Rules, Categories, and Combinators for \\ Syntax and Semantics}

\author{J{\o}rgen Villadsen}

\institute{Computer Science, Roskilde University \\[1ex] Building 42.1, DK-4000 Roskilde, Denmark \\[2ex] \email{\href{mailto:\jv}{\jv}}}

\maketitle


\begin{abstract}

We investigate the possibility of modelling the syntax and semantics of natural language by constraints, or rules, imposed by the multi-dimensional type theory Nabla.
The only multiplicity we explicitly consider is two, namely one dimension for the syntax and one dimension for the semantics, but the general perspective is important.
For example, issues of pragmatics could be handled as additional dimensions.

One of the main problems addressed is the rather complicated repertoire of operations that exists besides the notion of categories in traditional Montague grammar.
For the syntax we use a categorial grammar along the lines of Lambek.
For the semantics we use so-called lexical and logical combinators inspired by work in natural logic.
Nabla provides a concise interpretation and a sequent calculus as the basis for implementations.

\end{abstract}

\

\begin{small}
\noindent
~~\ldots~
Lambek originally presented his type logic as a calculus of \emph{syntactic} types.
Semantic interpretation of categorial deductions along the lines of the Curry-Howard correspondence was put on the categorial agenda in
J. van Benthem (1983)
\emph{The semantics of variety in categorial grammar}, Report 83-29*, Simon Fraser University, Canada.
This contribution made it clear how the categorial type logics realize Montague's Universal Grammar program ---
in fact, how they improve on Montague's own execution of that program in offering an integrated account of the composition of linguistic meaning \emph{and} form.
Montague's adoption of a categorial syntax does not go far beyond notation:
he was not interested in offering a principled theory of allowable `syntactic operations' going with the category formalism.

\hfill * Revised version in \cite{Buszkowski+88}

\

\noindent
M. Moortgat (1997) \emph{Categorial Type Logics}, in J. van Benthem \&\ A. ter Meulen (eds.) \emph{Handbook of Logic and Language}, Elsevier.

\

\vfill

\noindent
Full paper of short presentation at the International Workshop on \keywords\ (\subject).
This research was partly sponsored by the IT University of Copenhagen
and the CONTROL project: CONstraint based Tools for RObust Language processing
\,\href{http://control.ruc.dk}{\texttt{http://control.ruc.dk}}
\end{small}

\makeatletter


\def\infon{\gdef\@inffactor{1}}

\def\infhide{\gdef\@inffactor{0}}

\infon

\newdimen\premisskip	
\newdimen\ursep		
\newdimen\overshoot	
\newdimen\infthick	

\newbox\pp@box		
\newbox\p@box		
\newbox\f@box		
\newbox\n@box		

\newif\if@first@infere	
\newif\if@first@and	

\newdimen\p@skip	
\newdimen\p@width	

\newdimen\pp@skip	
\newdimen\pp@end	

\premisskip	1.0em
\ursep		0.3em
\overshoot	0.3em
\infthick	0.4pt

\def\d@tmp{\dimen0}
\def\n@skip{\dimen1}
\def\f@skip{\dimen2}
\def\l@skip{\dimen3}
\def\l@width{\dimen4}

\def\@infere{
   \if@first@infere
      \@first@inferefalse
      \setbox\p@box\box\f@box
      \p@skip=0pt
      \p@width\wd\p@box
   \else
      \f@skip\p@width
      \advance\f@skip by-\wd\f@box
      \divide\f@skip by\tw@
      \advance\f@skip by\p@skip
      \ifdim\f@skip>0pt
         \n@skip=0pt
      \else
         \n@skip=-\f@skip
         \f@skip=0pt
      \fi
      \d@tmp\n@skip
      \advance\d@tmp by\p@skip
      \l@skip\mindim{\f@skip}{\d@tmp}
      \l@width\p@skip
      \advance\l@width by \p@width
      \d@tmp\f@skip
      \advance\d@tmp by \wd\f@box
      \l@width\maxdim{\l@width}{\d@tmp}
      \advance\l@width by -\l@skip
      \p@skip\f@skip
      \p@width\wd\f@box
      \setbox\p@box
      \vbox
         {\hbox{\kern\n@skip\box\p@box}
          \nointerlineskip
          \hbox{\kern\l@skip\vrule width\l@width
\raisebox{-0.5ex}[\@inffactor\infthick\infon][0pt]{\kern\ursep\box\n@box}}
          \nointerlineskip
          \hbox{\kern\f@skip\box\f@box}
         }
   \fi
}
\def\mindim#1#2{\ifdim#1<#2#1\else#2\fi}
\def\maxdim#1#2{\ifdim#1>#2#1\else#2\fi}

\def\and@proof{
   \if@first@and
      \@first@andfalse
      \pp@skip\p@skip
      \pp@end\p@skip
      \advance\pp@end by \p@width
      \setbox\pp@box\box\p@box
   \else
      \pp@end\wd\pp@box
      \advance\pp@end by\premisskip
      \advance\pp@end by\p@skip
      \advance\pp@end by\p@width
      \setbox\pp@box\hbox{\vbox{\box\pp@box}\kern\premisskip\box\p@box}
   \fi}

\def\open@f@box{\setbox\f@box\vbox\bgroup\hbox\bgroup\kern\overshoot$\strut}
\def\close@f@box{$\kern\overshoot\egroup\egroup}

\def\open@p@box{\setbox\p@box
                \vbox
                   \bgroup
                      \b@infere}
\def\close@p@box{     \e@infere
                      \global\dimen0\p@skip
                      \global\dimen1\p@width
                   \egroup
                \p@skip\dimen0
                \p@width\dimen1
}

\def\b@infere{\@first@inferetrue\open@f@box}
\def\m@infere[#1]{\close@f@box\@infere\setbox\n@box\hbox{#1}\open@f@box}
\def\e@infere{\close@f@box\@infere\box\p@box}

\def\b@and{\close@f@box\@first@andtrue\open@p@box}
\def\m@and{\close@p@box\and@proof\open@p@box}
\def\e@and[#1]{\close@p@box\and@proof
               \@first@inferefalse
               \setbox\p@box\box\pp@box
               \p@skip\pp@skip
               \p@width\pp@end
               \advance\p@width by-\p@skip
               \setbox\n@box\hbox{#1}
               \open@f@box}

\def\@@infere{\@ifnextchar [{\m@infere}{\m@infere[]}}
\def\@@and{\@ifnextchar [{\e@and}{\e@and[]}}

\newenvironment{natproof}{
   \bgroup
       \let\\=\@@infere
       \let\Lproof=\b@and
       \let\ANDproof=\m@and
       \let\Rproof=\@@and
       \b@infere
}{     \e@infere
   \egroup
}

\newenvironment{displayproof}{
    \begin{displaymath}
        \begin{natproof}
}{      \end{natproof}
    \end{displaymath}
}

\makeatother

\section{Introduction}

We investigate the possibility of modelling the syntax and semantics of natural language
by constraints, or rules, imposed by the multi-dimensional type theory Nabla \cite{Villadsen95}.
The only multiplicity we explicitly consider here is two,
namely one dimension for the syntax and one dimension for the semantics, but we find the general perspective to be important.
For example, issues of pragmatics could be handled as additional dimensions
by taking into account direct references to language users and, possibly, other elements of the situation in which expressions are used.
We note that it is possible to combine many dimensions into a single dimension using Cartesian products.
Hence there is no theoretical difference between a one-dimensional type theory and a multi-dimensional type theory.
However, we think that in practice the gain can be substantial.

Nabla is a linguistic system based on categorial grammars \cite{Buszkowski+88}
and with so-called lexical and logical combinators \cite{Villadsen97-CRA} inspired by work in natural logic \cite{Sanchez91}.
The original goal was to provide a framework in which to do reasoning involving propositional attitudes
like knowledge and beliefs \cite{Villadsen01:LACL,Villadsen04:MATES}.

\opt{x}{

\subsection{Background}

In computational linguistics work on Nabla \cite{Villadsen01:LACL} has previously focussed entirely on the logical semantics of propositional attitudes 
replacing the classical higher order (intensional) logic \cite{Montague73*} to an
inconsistency-tolerant, or paraconsistent, (extensional) logic \cite{Villadsen04:AISC}
by introducting some kind of partiality, or indeterminacy, cf.~\cite{Muskens95}.
These ideas have also found applications outside natural language semantics as such, in particular in advanced databases and multi-agent systems \cite{Villadsen02:FQAS,Villadsen04:MATES}.

\subsection{Arguments}

In Nabla we can specify a grammar, that is, a definition of a set of
well-formed expressions.
Given the grammar, we can also specify a logic in Nabla.
The logic defines the notion of a correct argument. We use the mark $\surd$
for correct arguments and the mark $\div$ for incorrect arguments:

\begin{displayproof}
\Lproof
\M{John is a man.}
\ANDproof
\M{Victoria is a woman.}
\ANDproof
\M{John loves Victoria.}
\Rproof[~~~$\surd$]
\M{John loves a woman. \hspace{\premisskip} A man loves Victoria.}
\end{displayproof}

\begin{displayproof}
\M{John loves Victoria.}
\\[~~~$\div$]
\M{John loves a woman. \hspace{\premisskip} A man loves Victoria.}
\end{displayproof}

\noindent
The sentences above the line are the premises of the argument; the sentences
below the line are the conclusions of the argument.
By allowing multiple conclusions we obtain a nice symmetry around the line.

Note that the mark indicates what the logic says about the correctness
--- it does not say what is possible or not possible to infer
in any particular situation. For instance, if one do not understand
English or just do not understand the single word `is' (one might take it
to be synonymous to `hates') then it might be more appropriate to take
the first argument to be incorrect; or if one presuppose knowledge about
the conventions for male and female names then it might be more appropriate
to take the second argument to be correct.

To sum up: in Nabla we specify both a grammar and a logic;
these defines the set of all arguments (and hence the set of all sentences
and other subexpressions) and the set of correct arguments (and hence the
set of incorrect arguments is the remaining arguments).

\subsection{Formulas}

The logic for the formulas is here first order logic, also known as predicate logic \cite{vanBenthem91}.
The meaning of the first argument is the following formula:
$$
M J \land W V \land L J V \Rightarrow
  \exists y (W y \land L J y) \land \exists x (M x \land L x V)
$$
We use a rather compact notation.
We use lowercase letters for variables and uppercase letters for constants
(both for ordinary constants like $J$ and $V$ for `John' and `Victoria'
and for predicate constants like $M$ for `man', $W$ for `woman' and
$L$ for `love').
Note that conjunction $\land$ has higher priority than implication
$\Rightarrow$ and that the quantifier $\exists$ has even higher priority
(hence we need the parentheses to get the larger scope).

\subsection{Strings}

Traditionally, the map from arguments to formulas would consist of a
map from the individual sentences of the argument to formulas and a
procedure describing the assembling of the final formula from the
separate formulas.
The map from the sentences would have to deal with various inflections
and possibly minor items like punctuation and rules of capitalization.

Consider again the argument:

\begin{displayproof}
\Lproof
\M{John is a man.}
\ANDproof
\M{Victoria is a woman.}
\ANDproof
\M{John loves Victoria.}
\Rproof[~~~$\surd$]
\M{John loves a woman. \hspace{\premisskip} A man loves Victoria.}
\end{displayproof}

\noindent
In Nabla we make a single pass over the argument to obtain a string,
which is a sequence of tokens
(a token is to be thought of as a unit representing a word or
a phrase):
\begin{quote}
\T{John be a man also Victoria be a woman also John love Victoria so} \\
\T{John love a woman also a man love Victoria}
\end{quote}
Note the tokens \T{so} and \T{also} as well as the changes to the verbs
(the person and tense information is discarded since we only consider
present tense, third person).
We emphasize that the map from arguments to string is a simple bijection.
Only quite trivial manipulations are allowed and the overall
word-order must be unchanged.

\subsection{Combinators}

We provide a brief introduction to combinators and the $\lambda$-calculus
\cite{Schoenfinkel24,Stenlund72,Hindley+86}.
The combinators and the $\lambda$-calculus can be either typed or untyped \cite{Barendregt84};
we only consider the typed variant here as it is used to extend classical first order logic to higher order logic \cite{vanBenthem91}.

By $f a$ we mean the application of a function $f$ to an argument $a$.
It is possible to consider multiple arguments, but we prefer to regard
$f a b$ as $(f a) b$ and so on
(also known as currying, named after Curry though it was invented by Sch\"{o}nfinkel \cite{Schoenfinkel24,Stenlund72,Hindley+86}).

A combinator, say \C{x} or \C{y}, can manipulate the arguments:
$$
\C{x} f g \leadsto g f ~~~~~
\C{y} a b c \leadsto c b b
$$
The manipulations are swap ($f$ and $g$), deletion ($a$),
duplication ($b$), and permutation ($c$).
We can define the combinators using the so-called $\lambda$-abstraction:
$$
\C{x} \equiv \lambda a b (b a) ~~~~~
\C{y} \equiv \lambda a b c (c b b)
$$
Hence for example (the numbers are treated as constants):
$$
\C{x} 1 (\C{x} 2 \C{y} (\C{x} 3 4) 5) \leadsto
\C{x} 2 \C{y} (\C{x} 3 4) 5 1 \leadsto
\C{y} 2 (\C{x} 3 4) 5 1 \leadsto
5 (\C{x} 3 4) (\C{x} 3 4) 1 \leadsto
5 (4 3) (4 3) 1
$$
The $\lambda$-abstraction binds the variables (they were free before).
We call a combinator pure if it is defined without constants.
We always use uppercase letters for constants and lowercase letters for
variables. The combinators \C{x} and \C{y} are pure.

The following combinator \C{send} is not pure ($R$ is a constant):
$$
\C{send} \equiv \lambda a b c (R c b a)
$$
The combinator \C{send} means `sends \ldots\ to' and $R$ means `receives \ldots\ from'
(also possible to use the combinator \C{receive} and the constant $S$).
For example as in `Alice sends the box to Charlie' or `Charlie receives the box from Alice'.

We use the $\lambda$-calculus with the following rules
(observe that we write $f a$ rather than $f(a)$ for the application of
a function $f$ to an argument $a$):

\begin{itemize}
\item
$\alpha$-conversion
($y$ not free in $\alpha$ and $y$ free for $x$ in $\alpha$):
$$
\lambda x \alpha \leadsto \lambda y \alpha [x := y]
$$

\item
$\beta$-reduction
($\beta$ must be free for $x$ in $\alpha$):
$$
(\lambda x \alpha) \beta \leadsto \alpha [x := \beta]
$$

\item
$\eta$-reduction ($x$ not free in $\alpha$):
$$
\lambda x (\alpha x) \leadsto \alpha
$$

\end{itemize}
We use $\leadsto_{\lambda}$ for evaluation using these three rules
($\lambda$-conversion).

We use the (typed) $\lambda$-calculus \cite{vanBenthem91} in formulas (and combinator definitions).
The higher order logic in Montague grammar is also based on the $\lambda$-calculus,
but the usual rules of $\lambda$-conversion do not hold unrestricted,
due to the intensionality present \cite{Muskens95}.

Our use of combinators is inspired by work in natural logic
\cite{Purdy91-NDJFL,Purdy92-NDJFL,Sanchez91,Villadsen97-CRA}
and differs from previous uses in computer science, mathematical logic and natural language semantics
\cite{Curien86,Curry+58,Curry+72,Simons89-NDJFL,Steedman88*}.

\subsection{Type Language and Type Interpretation}

The basic ideas is closely related to the type theory by Morrill \cite{Morrill94}.
We take a type theory to consist of a type language and a type interpretation.

A type language $\cT$ is given by a set of basic types $\cT_{0} \subseteq \cT$
and a set of rules of type construction.
There is a rule of type construction for each type constructor.
Each type constructor makes a type out of subtypes.

A type interpretation consists of an interpretation function $\INT{\cdot}$
with respect to a universe.
A universe is a set of objects.
A subset of the universe is called a category (for example the empty
category and the universal category).
The interpretation function maps types to categories.
We may call a type a category name (or even just a category, and
the interpretation of the type for the category content).

A universe together with a type interpretation for basic types
$\INT{A}$ ($A \in \cT_{0}$) is a model.
The type interpretation for arbitrary types is defined compositionally
--- that is, the type interpretation is a composition
of the subtypes interpretations
(we have to stay within the universe, of course).
Hence we extend a basic type interpretation
$\INT{A}$ ($A \in \cT_{0}$)
to a type interpretation
$\INT{A}$ ($A \in \cT$).
It is essential that we do not think of objects as atomic.
They can have components; hence we get a multi-dimensional type theory.
Let $n$ be the number of dimensions.

Each type is interpreted as a category ---
the members hereof are called inhabitants.
An inhabitation is a category for each type.
An inhabitation extends another inhabitation if and only if (iff) for each type,
the category of the former includes the category of the latter.

With respect to the type interpretation an initial inhabitation determines
a final inhabitation as its minimal extension satisfying the interpretation
of types (we assume that such a minimal extension exists).
Note that the interpretation of types is a precise definition of
the inhabitants of a type based on the inhabitant of its subtypes.

\subsection{Theory of Inhabitation and Theory of Formation}

We emphasize that an inhabitation is not (just) a basic type interpretation
(this holds for initial inhabitations too).

An arrow is a component-wise operation on objects labelled by types.
An inhabitation satisfies an arrow iff it is closed under the arrow.
A theory of inhabitation is a set of arrows.
An inhabitation satisfies a theory of inhabitation iff
it satisfies every arrow in the theory of inhabitations.
An initial inhabitation together with a theory of inhabitation
determine a final inhabitation which is the minimal extension of
the initial inhabitation satisfying the theory of inhabitation.

In order to represent objects and arrow (and inhabitations and
theories of inhabitations) we introduce representation languages
(let $a_i$ range over terms of the representation language for dimension $i$).
An entry is a sequence of terms and a type, written as
$a_1 - \ldots - a_n : A$
(where $n$ is the number of dimensions).
A formation is a set of entries.
A sequent or a statement of formation is a configuration and
an entry, written as
$\Delta ~\succ~ a_1 - \ldots - a_n : A$
(where the left side contains the antecedents and the right side contains
the succedent).
A configuration is a finite set of sequences of variable de\-clarations
$\{ x^1_1 - \ldots - x^n_1 : A_1, \ldots, x^1_m - \ldots - x^n_m : A_m \}$.
A statement of formation gives a formation as all instantiations of variables.
A theory of formation is a set of statements of formation.
An initial formation plus a theory of formation
give a final formation
in the same way as
an initial inhabitation plus a theory of inhabitation
give a final inhabitation.

We provide a theory of formation by a set of rules of formation,
which defines the theory of formation inductively.

\subsection{Nabla}

The main task of Nabla is then to define a total set of strings and for each string a set of formulas.
If the set of formulas for a string is empty it indicates that the string
does not map to an argument.
If the set of formulas for a string has more than one member then it shows
that the string maps to an ambiguous argument
(one or more of the sentences are ambiguous).

In Nabla the grammar is completely given by a lexicon (there are no rules specific for the particular fragment of natural language).
The lexicon has a set of entries for each token; the set of tokens is
called the vocabulary.
The grammar determines the string / formula association and the logic determines
the validity of the formula.
Besides the grammar and the logic we also need a tokenizer which is a
quite simple device that turns arguments into strings.

\subsection{Overview}

In sections~2, 3, and 4 we present the rules, the categories, and the combinators, respectively.

Section~5 provides examples and section~6 concludes.

}

\section{The Rules}

We define a multi-dimensional type theory with the two dimensions: syntax and semantics.
We use a kind of the so-called Lambek calculus with the two type constructors $\FS$ and $\BS$,
which are right- and left-looking functors \cite{Lambek58-AMM,Moortgat88,Morrill94}.

We assume a set of basic types $\cT_{0}$, where $\BULLET \in \cT_{0}$
is interpreted as truth values.
The set of types $\cT$ is the smallest set of expressions containing
$\cT_{0}$ such that if $A,B \in \cT$ then $A \FS B, B \BS A \in \cT$.

A structure consists of a vocabulary and a set of bases
$\cS \equiv \langle \cV, \cB \rangle$, where
$\cV$ is finite and
$\cB(A) \ne \emptyset$ for all $A \in \cT_{0}$.

We define three auxiliary functions on types
(the first for the syntactic dimension and the second for
the semantic dimension; symbol $\doteq$ is used for such ``mathematical'' definitions,
in contrast with $\equiv$ for literal definitions):
$$
\begin{array}{l}
\SYN{A} \doteq \cV^{+}, A \in \cT
\\[2ex]
\SEM{A} \doteq \cB(A), A \in \cT_{0}
\\
\SEM{A \FS B} \doteq \SEM{B \BS A} \doteq \SEM{B} \to \SEM{A}
\\[2ex]
\SYNSEM{A} \doteq \SYN{A} \times \SEM{A}
\end{array}
$$
By $\cV^{+}$ we mean the set of (non-empty) sequences of elements from $\cV$
(such sequences correspond to strings and for the sake of simplicity
we call them strings).

The universe is $\bigcup_{A \in \cT} \SYNSEM{A}$ (which depends only
on the structure $\cS$).

With respect to $\cS$ we extend a basic type interpretation
$\INT{A} \subseteq \SYNSEM{A}$ ($A \in \cT_{0}$)
to a type interpretation
$\INT{A} \subseteq \SYNSEM{A}$ ($A \in \cT$)
as follows (the concatenation of the strings $x$ and $x'$ is written $x \CONCAT x'$):
$$
\begin{array}{l}
\INT{A \FS B} \doteq \{~ \langle x,y \rangle ~|~ \M{
for all $x', y'$,
if $\langle x',y' \rangle \in \INT{B}$
then $\langle x \CONCAT x',y y' \rangle \in \INT{A}$
} ~\}
\\[1ex]
\INT{B \BS A} \doteq \{~ \langle x,y \rangle ~|~ \M{
for all $x', y'$,
if $\langle x',y' \rangle \in \INT{B}$
then $\langle x' \CONCAT x,y y' \rangle \in \INT{A}$
} ~\}
\end{array}
$$

We use a so-called sequent calculus \cite{Prawitz65} with an explicit semantic dimension and an implicit
syntactic dimension.
The implicit syntactic dimension means that the antecedents form a sequence
rather than a set and that the syntactic component for the succedent
is the concatenation of the strings for the antecedents.
It should be observed that all rules work unrestricted on the semantic
component from the premises to the conclusion.
We refer to the resulting sequent calculus as the Nabla calculus.

We use $\Gamma$ (and $\Delta$) for sequences of categories $A_{1}\ldots A_{n}$ ($n>0$).
The rules have sequents of the form $\Gamma ~\succ~ A$.
The sequent means that if $a_{1}, \ldots, a_{n}$ are strings
of categories $A_{1}, \ldots, A_{n}$, respectively, then the string
that consists of the concatenation of the strings $a_{1}, \ldots, a_{n}$
is a string of category $A$.
Hence the sequent $A ~\succ~ A$ is valid for any category $A$.

Rules are displayed starting with the conclusion and the premises indented below.
There are two rules for $\FS$ (a left and a right rule) and two rules for $\BS$ too.
The left rules specify how to introduce a $\FS$ or a $\BS$ at the left
side of the sequent symbol $\succ$, and vice versa for the right rules
(observe that the introduction is in the conclusion and not in the
premises).
The reason why we display the rules in this way is that sequents tend to
get very long, often as long as a whole line, and hence the more usual tree format would be problematic.
Also the conclusion is usually longer than each of the premises.

We note that only the right rule of $\lambda$ (where $\alpha \leadsto_{\lambda} \alpha'$ is $\lambda$-conversion) is possible,
since only variables are allowed on the left side of the sequent symbol.

\medskip

\noindent
\LINE{0}{x : A}{x : A}{=}
\\[2ex]
\LINE{0}{\Delta}{\alpha' : A}
  {$\alpha \leadsto_{\lambda} \alpha'$ ~~~ $\lambda$} \\[1ex]
\LINE{1}{\Delta}{\alpha : A}{}
\\[2ex]
\LINE{0}{\Delta[\Gamma]}{\beta[x \mapsto \alpha] : B}{Cut} \\[1ex]
\LINE{1}{\Gamma}{\alpha : A}{} \\[1ex]
\LINE{1}{\Delta[x : A]}{\beta : B}{}
\\[2ex]
\LINE{0}{\Delta[\Gamma ~~ z : B \BS A]}
        {\gamma[x \mapsto (z ~ \beta)] : C}{\BS L} \\[1ex]
\LINE{1}{\Gamma}{\beta : B}{} \\[1ex]
\LINE{1}{\Delta[x : A]}{\gamma : C}{}
\\[2ex]
\LINE{0}{\Gamma}{\lambda y \alpha : B \BS A}{\BS R} \\[1ex]
\LINE{1}{y : B ~~ \Gamma}{\alpha : A}{}
\\[2ex]
\LINE{0}{\Delta[z : A \FS B ~~ \Gamma]}
        {\gamma[x \mapsto (z ~ \beta)] : C}{\FS L} \\[1ex]
\LINE{1}{\Gamma}{\beta : B}{} \\[1ex]
\LINE{1}{\Delta[x : A]}{\gamma : C}{}
\\[2ex]
\LINE{0}{\Gamma}{\lambda y \alpha : A \FS B}{\FS R} \\[1ex]
\LINE{1}{\Gamma ~~ y : B}{\alpha : A}{}



\opt{x}{

\subsection{Comments}

The order of the premises does not matter, but we adopt the convention
that the minor premises (the premises that ``trigger'' the introduction of
$\FS$ or $\BS$) come first and the major premises (the premises that
``circumscribes'' the introduction of $\FS$ or $\BS$) come second.

The rule \FS R is to be understood as follows:
if we prove that (the syntactic components for the types in) $\Gamma$
with (the syntactic component for the type) $B$ to the right
yield (the syntactic component for the type) $A$,
then we conclude that (\ldots) $\Gamma$ (alone) yields (\dots) $A \FS B$;
furthermore if the variable $y$ represents
(the semantic component for the type) $B$
and the term $\alpha$ represents
(the semantic component for the type) $A$,
then the $\lambda$-abstraction $\lambda y \alpha$ represents
(\ldots) $A \FS B$
(we do not care about the semantic components for the types in $\Gamma$
since these are being taken care of in $\alpha$).

In the same manner the rule \FS L is to be understood as follows:
if we prove that $\Gamma$ yields $B$ and also prove that
$\Delta$ with $A$ inserted yields $C$,
then we conclude that $\Delta$ with $A \FS B$ and $\Gamma$ (in that order)
inserted (at the same spot as in the premise) yields $C$;
furthermore if the term $\beta$ represents $B$ and
the term $\gamma$ represents $C$
(under the assumption that the variable $x$ represents $A$),
then $\gamma$ with the application $(z ~ \beta)$ substituted for
all free occurrences of the variable $x$ represents $C$
(under the assumption that the variable $z$ represents $A \FS B$).

}

\section{The Categories}

As basic categories for the lexicon we have $N$, $G$, $S$ and the
top category \BULLET\ corresponding to the whole argument
(do not confuse the basic category $N$ with the constant $N$ for `Nick'
and so on).
Roughly we have that $N$ corresponds to ``names'' (proper nouns),
$G$ corresponds to ``groups'' (common nouns) and $S$ to ``sentences''
(discourses).
Consider the following lexical category assignments:

\bigskip

\noindent
\CAT{John~~Nick~~Gloria~~Victoria}{N} \\[.4ex]
\CAT{run~~dance~~smile}{N \BS S} \\[.4ex]
\CAT{find~~love}{(N \BS S) \FS N} \\[.4ex]
\CAT{man~~woman~~thief~~unicorn}{G} \\[.4ex]
\CAT{popular~~quick}{G \FS G} \\[.4ex]
\CAT{be}{(N \BS S) \FS N} \\[.4ex]
\CAT{be}{(N \BS S) \FS (G \FS G)} \\[.4ex]
\CAT{a~~every}{(S \FS (N \BS S)) \FS G ~~~ ((S \FS N) \BS S) \FS G} \\[.4ex]
\CAT{not}{(N \BS S) / (N \BS S)} \\[.4ex]
\CAT{nix}{S \FS S} \\[.4ex]
\CAT{and~~or}{S \BS (S \FS S)} \\[.4ex]
\CAT{and~~or}{(N \BS S) \BS ((N \BS S) \FS (N \BS S)) ~~~ (G \FS G) \BS ((G \FS G) \FS (G \FS G))} \\[.4ex]
\CAT{ok}{S} \\[.4ex]
\CAT{also}{S \BS (S \FS S)} \\[.4ex]
\CAT{so}{S \BS (\BULLET \FS S)}

\bigskip

\opt{x}{

\subsection{Comments}

The order of the tokens is the same as for the lexical combinators to come.

Together the lexical category assignments and the lexical combinator
definitions constitute a set of lexical entries.
A lexicon consists of a skeleton and set of lexical entries.
For the lexicon the skeleton is just the three basic categories
$N$, $G$ and $S$ (we omit the top category, which is always \BULLET).

Note that it is not a mistake that there is no $'$ for, say, the token \T{be}
(compared with the combinators \C{be} and \C{be'}).
There simply is just one token (with two meanings).

}

\section{The Combinators}

We introduce the following so-called logical combinators \cite{Stenlund72}:
$$
\begin{array}{cll@{~~~~~~}l}
\dot{\C{Q}}
& \equiv & \lambda x y (x = y) & \rm Equality \\[.4ex]
\dot{\C{N}}
& \equiv & \lambda a (\lnot a) & \rm Negation \\[.4ex]
\dot{\C{C}}
& \equiv & \lambda a b (a \land b) & \rm Conjunction \\[.4ex]
\dot{\C{D}}
& \equiv & \lambda a b (a \lor b) & \rm Disjunction \\[.4ex]
\dot{\C{O}}
& \equiv & \lambda t u \exists x (t x \land u x) & \rm Overlap \\[.4ex]
\dot{\C{I}}
& \equiv & \lambda t u \forall x (t x \Rightarrow u x) & \rm Inclusion \\[.4ex]
\dot{\C{T}}
& \equiv & \top & \rm Triviality \\[.4ex]
\dot{\C{P}}
& \equiv & \lambda a b (a \Rightarrow b) & \rm Preservation
\end{array}
$$
After having introduced the logical combinators we introduce the so-called
lexical combinators.
There is one or more combinator for each token in the vocabulary, for example
the combinator \C{John} for the token \T{John}, \C{be} and \C{be'} for
\T{be} and so on
(tokens and combinators are always spelled exactly the same way except
for the $'$ (possibly repeated) at the end).

In order to display the lexicon more compactly we introduce two
place-holders (or ``holes'') for combinators and constants, respectively.
$\BIGCIRC$ is place-holder for logical combinators (if any) and
$\CIRC$ is place-holder for (ordinary and predicate) constant (if any);
the combinators and constants to be inserted are shown after the $|$ as
in the following lexicon:

\bigskip

\noindent
\DEFX{John~~Nick~~Gloria~~Victoria}{\CIRC}{J~~N~~G~~V} \\[.4ex]
\DEFX{run~~dance~~smile}
  {\lambda x (\CIRC x)}
  {R~~D~~S} \\[.4ex]
\DEFX{find~~love}
  {\lambda y x (\CIRC x y)}
  {F~~L} \\[.4ex]
\DEFX{man~~woman~~thief~~unicorn}
  {\lambda x (\CIRC x)}
  {M~~W~~T~~U} \\[.4ex]
\DEFXX{popular~~quick}
  {\lambda t x (\BIGCIRC (\CIRC x) (t x))}
  {\dot{\C{C}}}{P~~Q} \\[.4ex]
\DEFX{be}{\lambda y x (\BIGCIRC x y)}{\dot{\C{Q}}} \\[.4ex]
\DEFX{be'}{\lambda f x (f \lambda y (\BIGCIRC x y) x)}{\dot{\C{Q}}} \\[.4ex]
\DEFX{a~~every}
  {\lambda t u (\BIGCIRC t u)}
  {\dot{\C{O}}~~\dot{\C{I}}} \\[.4ex]
\DEFX{not}{\lambda t x (\BIGCIRC (t x))}{\C{N}} \\[.4ex]
\DEFX{nix}{\lambda a (\BIGCIRC a)}{\C{N}} \\[.4ex]
\DEFX{and~~or}{\lambda a b (\BIGCIRC a b)}{\C{C}~~\C{D}} \\[.4ex]
\DEFX{and'~~or'}{\lambda t u x (\BIGCIRC (t x) (u x))}{\C{C}~~\C{D}} \\[.4ex]
\DEFX{ok}{\BIGCIRC}{\dot{\C{T}}} \\[.4ex]
\DEFX{also}{\lambda a b (\BIGCIRC a b)}{\dot{\C{C}}} \\[.4ex]
\DEFX{so}{\lambda a b (\BIGCIRC a b)}{\dot{\C{P}}}

\bigskip

\opt{x}{

\subsection{Comments}

It might be possible to display the lexicon in an even more compact way
by avoiding the remaining repetitions (for the token \T{be}), but we
have not found it worthwhile.

We have put dots above the logical combinators in order to distinguish them
from the more advanced logical combinators previously used \cite{Villadsen01:LACL}
(the first four are the same; we will not go into details about the remaining *-marked combinator
which also uses the special predicates $E$ for existence and $I$ for integrity in a paraconsistent logic):
$$
\begin{array}{cll@{~~~~~~}l}
\C{Q} & \equiv & \lambda x y (x = y) & \rm Equality \\[.4ex]
\C{N} & \equiv & \lambda a (\lnot a) & \rm Negation \\[.4ex]
\C{C} & \equiv & \lambda a b (a \land b) & \rm Conjunction \\[.4ex]
\C{D} & \equiv & \lambda a b (a \lor b) & \rm Disjunction \\[.4ex]
\C{E} & \equiv &
  \lambda i t \exists x (E i x \land t x) &
  \rm Existentiality* \\[.4ex]
\C{U} & \equiv &
  \lambda i t \forall x (E i x \Rightarrow t x) &
  \rm Universality* \\[.4ex]
\C{O} & \equiv &
  \lambda i t u \exists x (E i x \land (t x \land u x)) &
  \rm Overlap* \\[.4ex]
\C{I} & \equiv &
  \lambda i t u \forall x (E i x \Rightarrow (t x \Rightarrow u x)) &
  \rm Inclusion* \\[.4ex]
\C{T} & \equiv & \lambda i (I i) & \rm Triviality* \\[.4ex]
\C{F} & \equiv & \lambda i p (I i \land p i) & \rm Filtration* \\[.4ex]
\C{P} & \equiv & \lambda p q \forall i (p i \Rightarrow q i) & \rm Preservation*
\end{array}
$$
Note that even though we use the $\lambda$-calculus of higher order logic we have not used
any higher-order quantifications.

The overlap combinator takes two sets and test for overlap (analogously
for the inclusion combinator).
The triviality combinator is used in case of no premises or
no conclusions in an argument.
The preservation combinator is used between the premises and
the conclusions in an argument.

Some of the combinators are discussed elsewhere, cf.\ \cite[page~270]{Hindley+86},
in particular the `restricted generality' combinator $\Xi$ corresponding
to our logical combinator $\dot{\C{I}}$ (see also \cite{Curry+58}),
but usually using the untyped $\lambda$-calculus with a definition of so-called
canonical terms in order to avoid a paradox discovered by Curry.

Let us return to the formula we considered in the introduction:
$$
M J \land W V \land L J V \Rightarrow
  \exists y (W y \land L J y) \land \exists x (M x \land L x V)
$$
Using the logical combinators we obtain the formula:
\begin{quote}
$
\C{\dot{P}} ~
\\\mbox{}
~~~~~
  (\C{\dot{C}} ~ (M J) ~ (\C{\dot{C}} ~ (W V) ~ (L J V))) ~
\\\mbox{}
~~~~~
  (\C{\dot{C}} ~
    (\C{\dot{O}} ~ \lambda y (W y) ~ \lambda y (L J y)) ~
    (\C{\dot{O}} ~ \lambda x (M x) ~ \lambda x (L x V)))
$
\end{quote}
Due to the $\eta$-rule in the $\lambda$-calculus
there is no difference between $W$ and $\lambda y (W y)$,
between $M$ and $\lambda x (M x)$, or between $L J$ and $\lambda y (L J y)$,
but there is no immediate alternative for $\lambda x (L x V)$.
We do not have to list
the types of constants since either the type of a constant is \E\
or the type can be determined by the types of its arguments.

At a first glance it may appear as though the use of combinators just makes
the formula look more complicated, but we have really added much more structure to
the formula.
Also, we are so used to the usual formulas of predicate logic that any change
is problematic. As soon as we leave the lexicon and turn to string / formula associations,
the use of logical combinators is easier to accept.

Let us add even more structure to the formula by using the equality
combinator (it is triggered by the word `is' in the two first sentences):
\begin{quote}
$
\C{\dot{P}} ~
\\\mbox{}
~~~~~
  (\C{\dot{C}} ~
    (\C{\dot{O}} ~ \lambda x (M x) ~ \lambda x (\C{\dot{Q}} x J)) ~
    (\C{\dot{C}} ~
      (\C{\dot{O}} ~ \lambda x (W x) ~ \lambda x (\C{\dot{Q}} x V)) ~
      (L J V))) ~
\\\mbox{}
~~~~~
  (\C{\dot{C}} ~
    (\C{\dot{O}} ~ \lambda y (W y) ~ \lambda y (L J y)) ~
    (\C{\dot{O}} ~ \lambda x (M x) ~ \lambda x (L x V)))
$
\end{quote}
Finally we would like to emphasize that it is not in any way a goal to get rid
of all variables although this is surely possible by introducing a series
of pure combinators, since the pure combinators
in general do not add any useful structure to the formula.
We think that the challenge is to find the best balance between
the use of combinators and the use of $\lambda$-abstractions.

Let us return to the previous formula with the logical combinators:
\begin{quote}
$
\C{\dot{P}} ~
\\\mbox{}
~~~~~
  (\C{\dot{C}} ~
    (\C{\dot{O}} ~ \lambda x (M x) ~ \lambda x (\C{\dot{Q}} x J)) ~
    (\C{\dot{C}} ~
      (\C{\dot{O}} ~ \lambda x (W x) ~ \lambda x (\C{\dot{Q}} x V)) ~
      (L J V))) ~
\\\mbox{}
~~~~~
  (\C{\dot{C}} ~
    (\C{\dot{O}} ~ \lambda y (W y) ~ \lambda y (L J y)) ~
    (\C{\dot{O}} ~ \lambda x (M x) ~ \lambda x (L x V)))
$
\end{quote}
Using the lexical combinators we obtain the formula:
\begin{quote}
$
\C{so} ~
\\\mbox{}
~~~~~
  (\C{also} ~
    (\C{a} ~ \C{man} ~ \lambda x (\C{be} ~ x ~ \C{John})) ~
\\\mbox{}
~~~~~
~~~~~
    (\C{also} ~
      (\C{a} ~ \C{woman} ~ \lambda x (\C{be} ~ x ~ \C{Victoria})) ~
      (\C{love} ~ \C{Victoria} ~ \C{John}))) ~
\\\mbox{}
~~~~~
  (\C{also} ~
    (\C{a} ~ \C{woman} ~ \lambda x (\C{love} ~ x ~ \C{John})) ~
    (\C{a} ~ \C{man} ~ (\C{love} ~ \C{Victoria})))
$
\end{quote}
We find this formula remarkably elegant.
What remains is the association with the original string:
\begin{quote}
\T{John be a man also Victoria be a woman also John love Victoria so} \\
\T{John love a woman also a man love Victoria}
\end{quote}
This is taken care of by the Nabla calculus. We now turn to some examples.

}

\section{Examples: Syntax and Semantics}

Consider the tiny argument (where $\surd$ indicates that the argument is correct):

\begin{displayproof}
\M{John is a popular man.}
\\[~~~$\surd$]
\M{John is popular.}
\end{displayproof}

\noindent
The lexical category assignments to tokens give us the following string / formula association
using the sequent calculus:
\begin{quote}
$
\T{John be a popular man so John be popular}
\\[1ex]
\LEAD
\C{so} ~
  (\C{a} ~ (\C{popular} ~ \C{man}) ~ \lambda x (\C{be} ~ x ~ \C{John})) ~
  (\C{be'} ~ \C{popular} ~ \C{John})
\\[1ex]
\LEAD
\C{\dot{P}} ~
  (\C{\dot{O}} ~ \lambda x (\C{\dot{C}} ~ (P x) ~ (M x)) ~
    \lambda x (\C{\dot{Q}} J x)) ~
  (\C{\dot{C}} ~ (P J) ~ (\C{\dot{Q}} J J))
\\[1ex]
\LEAD
P J \land M J \Rightarrow P J
$
\end{quote}
It is really an impressive undertaking, since not only does the order
of the combinators not match the order of the tokens, but there is also
no immediate clue in the string on how to get the structure of the
formula right (``the parentheses'').

As expected the resulting formula is valid.

\opt{x}{

\subsection{Step-by-Step Formula Extraction}

We consider the following tiny argument with one premise and no conclusion
(rather special, but good enough as an example):

\begin{displayproof}
\M{John smiles.}
\\[~~~$\surd$]
\end{displayproof}

\noindent
We show that the derivations for this argument yield a formula reducible
to $\top$ (and hence that the argument is a correct argument
as every argument with no conclusions is).
The argument corresponds to the following string:
\begin{quote}
\T{John smile so ok}
\end{quote}
The token \T{so} corresponds to the line in the argument
and the token \T{ok} corresponds to the omitted conclusions.
The string corresponds to the following sequent:
\begin{quote}
\LINE{0}{N ~~ N \BS S ~~ S \BS (\BULLET \FS S) ~~ S}
        {\BULLET}{}
\end{quote}
Note that \BULLET\ is the top category (arguments).
The other categories are given by the lexical category assignments.
By using the rules of the Nabla calculus we obtain the following derivation:

\

\LINE{0}{N ~~ N \BS S ~~ S \BS (\BULLET \FS S) ~~ S}
        {\BULLET}{\BS L}
\\[1.6ex]
\LINE{1}{N ~~ N \BS S}
        {S}{\BS L}
\\[1.6ex]
\LINE{2}{N}{N}{=}
\\[1.6ex]
\LINE{2}{S}{S}{=}
\\[1.6ex]
\LINE{1}{\BULLET \FS S ~~ S}
        {\BULLET}{\FS L}
\\[1.6ex]
\LINE{2}{S}{S}{=}
\\[1.6ex]
\LINE{2}{\BULLET}{\BULLET}{=}

\subsubsection{Step 1}

For simplicity we use numbers 1, 2, 3, \ldots\ as variables.
We start from the last line in the derivation, introduce the variables
1 and 2, and use the rule \FS L to get the term $3 ~ 2$
(the variable 3 is a fresh variable at the position where the $\FS$
is introduced):

\

\LINE{0}{3 ~~ 2}
        {3 ~ 2}{\FS L}
\\[1.6ex]
\LINE{1}{2}{2}{=}
\\[1.6ex]
\LINE{1}{1}{1}{=}

\

\subsubsection{Step 2}

We reuse the variable 1 and introduce the variable 4, and use the
rule \BS L to get the term $4 ~ 5$
(the variable 5 is a fresh variable at the position where the $\BS$
is introduced).
At last we use the rule \BS L to get the term $1 ~ (5 ~ 4) ~ 2$
(the variable 2 can be reused):

\

\LINE{0}{4 ~~ 5 ~~ 1 ~~ 2}
        {1 ~ (5 ~ 4) ~ 2}{\BS L}
\\[1.6ex]
\LINE{1}{4 ~~ 5}
        {5 ~ 4}{\BS L}
\\[1.6ex]
\LINE{2}{4}{4}{=}
\\[1.6ex]
\LINE{2}{1}{1}{=}
\\[1.6ex]
\LINE{1}{3 ~~ 2}
        {3 ~ 2}{\FS L}
\\[1.6ex]
\LINE{2}{2}{2}{=}
\\[1.6ex]
\LINE{2}{1}{1}{=}

\

\subsubsection{Step 3}

The tokens of the string correspond to the variables 4, 5, 1 and 2,
respectively, and the lexical combinators are inserted yielding the extracted formula.
Using the logical combinators the formula is then finally reduced to $\top$ as promised:

\smallskip

\begin{quote}
$
\T{John smile ok so}
\\[1ex]
\LEAD
\C{so} ~ (\C{smile} ~ \C{John}) ~ \C{ok}
\\[1ex]
\LEAD
\lambda a b (\C{\dot{P}} ~ a ~ b) ~ (\lambda x (S x) ~ J) ~ \C{\dot{T}}
\\[1ex]
\LEAD
\C{\dot{P}} ~ (S J) ~ \C{\dot{T}}
\\[1ex]
\LEAD
\lambda a b (a \Rightarrow b) ~ (S J) ~ \top
\\[1ex]
\LEAD
S J \Rightarrow \top
\\[1ex]
\LEAD
\top
$
\end{quote}
This completes the step-by-step example.

\subsection{Further Examples}

We first consider the argument using the string from the tokenizer:

\begin{displayproof}
\Lproof
\M{John runs.}
\ANDproof
\M{John is Nick.}
\Rproof[~~~$\surd$]
\M{Nick runs.}
\end{displayproof}
\vspace{.0ex}
\begin{quote}
\T{John run also John be Nick so Nick run}
\end{quote}
Here \T{John} has category $N$, \T{run} has category $N \BS S$,
\T{also} has category $S \BS (S \FS S)$ and so on.
The string has the top category \BULLET, since it is an argument.
By using the rules of the Nabla calculus we obtain the following derivation:

\bigskip

\LINE{0}{N ~~ N \BS S ~~ S \BS (S \FS S) ~~ N ~~ (N \BS S) \FS N ~~ N ~~
         S \BS (\BULLET \FS S) ~~ N ~~ N \BS S}
        {\BULLET}{\BS L}
\\[1.6ex]
\LINE{1}{N ~~ N \BS S ~~ S \BS (S \FS S) ~~ N ~~ (N \BS S) \FS N ~~ N}
        {S}{\BS L}
\\[1.6ex]
\LINE{2}{N ~~ N \BS S}
        {S}{\BS L}
\\[1.6ex]
\LINE{3}{N}{N}{=}
\\[1.6ex]
\LINE{3}{S}{S}{=}
\\[1.6ex]
\LINE{2}{S \FS S ~~ N ~~ (N \BS S) \FS N ~~ N}
        {S}{\FS L}
\\[1.6ex]
\LINE{3}{N ~~ (N \BS S) \FS N ~~ N}
        {S}{\FS L}
\\[1.6ex]
\LINE{4}{N}{N}{=}
\\[1.6ex]
\LINE{4}{N ~~ N \BS S}
        {S}{\BS L}
\\[1.6ex]
\LINE{5}{N}{N}{=}
\\[1.6ex]
\LINE{5}{S}{S}{=}
\\[1.6ex]
\LINE{3}{S}{S}{=}
\\[1.6ex]
\LINE{1}{\BULLET \FS S ~~ N ~~ N \BS S}
        {\BULLET}{\FS L}
\\[1.6ex]
\LINE{2}{N ~~ N \BS S}
        {S}{\BS L}
\\[1.6ex]
\LINE{3}{N}{N}{=}
\\[1.6ex]
\LINE{3}{S}{S}{=}
\\[1.6ex]
\LINE{2}{\BULLET}{\BULLET}{=}

\bigskip

\noindent
We extract the following formula for the derivation.
\begin{quote}
$
\T{John run also John be Nick so Nick run}
\\[1ex]
\LEAD
\C{so} ~ (\C{also} ~ (\C{run} ~ \C{John}) ~
  (\C{be} ~ \C{Nick} ~ \C{John})) ~ (\C{run} ~ \C{Nick})
\\[1ex]
\LEAD
\C{\dot{P}} ~ (\C{\dot{C}} ~ (R J) ~ (\C{\dot{Q}} J N)) ~ (R N)
\\[1ex]
\LEAD
R J \land J = N \Rightarrow R N
$
\end{quote}
Observe the reverse order of \C{John} and \C{Nick} in the formula with the
lexical combinators. All transitive verbs and the copula (token \T{be})
have the object before the subject in formulas with lexical combinators.
In the final formula the order is not reversed.

Only left rules were used in the derivation above.
The following argument requires a right rule due to the existential quantifier (token \T{a}) in the
object position of the copula (token \T{be} with two lexical combinators \C{be} and \C{be'}):

\begin{displayproof}
\M{John is a popular man.}
\\[~~~$\surd$]
\M{John is popular.}
\end{displayproof}
\vspace{.0ex}
\begin{quote}
\T{John be a popular man so John be popular}
\end{quote}

\bigskip

\LINEX{0}{N ~~ (N \BS S) \FS N ~~ ((S \FS N) \BS S) \FS G ~~ G \FS G ~~ G}
\\
\LINEY{S \BS (\BULLET \FS S) ~~ N ~~ (N \BS S) \FS (G \FS G) ~~ G \FS G}
      {\BULLET}{\BS L}
\\[1.6ex]
\LINE{1}{N ~~ (N \BS S) \FS N ~~ ((S \FS N) \BS S) \FS G ~~ G \FS G ~~ G}
        {S}{\FS L}
\\[1.6ex]
\LINE{2}{G}{G}{=}
\\[1.6ex]
\LINE{2}{N ~~ (N \BS S) \FS N ~~ ((S \FS N) \BS S) \FS G ~~ G}
        {S}{\FS L}
\\[1.6ex]
\LINE{3}{N ~~ (N \BS S) \FS N ~~ (S \FS N) \BS S}
        {S}{\BS L}
\\[1.6ex]
\LINE{4}{N ~~ (N \BS S) \FS N}
        {S \FS N}{\FS R}
\\[1.6ex]
\LINE{5}{N ~~ (N \BS S) \FS N ~~ N}
        {S}{\FS L}
\\[1.6ex]
\LINE{6}{N}{N}{=}
\\[1.6ex]
\LINE{6}{N ~~ N \BS S}
        {S}{\BS L}
\\[1.6ex]
\LINE{7}{N}{N}{=}
\\[1.6ex]
\LINE{7}{S}{S}{=}
\\[1.6ex]
\LINE{4}{S}{S}{=}
\\[1.6ex]
\LINE{3}{G}{G}{=}
\\[1.6ex]
\LINE{1}{\BULLET \FS S ~~ N ~~ (N \BS S) \FS (G \FS G) ~~ G \FS G}
        {\BULLET}{\FS L}
\\[1.6ex]
\LINE{2}{N ~~ (N \BS S) \FS (G \FS G) ~~ G \FS G}
        {S}{\FS L}
\\[1.6ex]
\LINE{3}{G \FS G}{G \FS G}{=}
\\[1.6ex]
\LINE{3}{N ~~ N \BS S}
        {S}{\BS L}
\\[1.6ex]
\LINE{4}{N}{N}{=}
\\[1.6ex]
\LINE{4}{S}{S}{=}
\\[1.6ex]
\LINE{2}{\BULLET}{\BULLET}{=}

\bigskip

\begin{quote}
$
\T{John be a popular man so John be popular}
\\[1ex]
\LEAD
\C{so} ~
  (\C{a} ~ (\C{popular} ~ \C{man}) ~ \lambda x (\C{be} ~ x ~ \C{John})) ~
  (\C{be'} ~ \C{popular} ~ \C{John})
\\[1ex]
\LEAD
\C{\dot{P}} ~
  (\C{\dot{O}} ~ \lambda x (\C{\dot{C}} ~ (P x) ~ (M x)) ~
    \lambda x (\C{\dot{Q}} J x)) ~
  (\C{\dot{C}} ~ (P J) ~ (\C{\dot{Q}} J J))
\\[1ex]
\LEAD
P J \land M J \Rightarrow P J
$
\end{quote}
Here the use of lexical and logical combinators is more substantial.

}

\section{Conclusions and Future Work}

The multi-dimensional type theory Nabla provides a concise interpretation and a sequent calculus as the basis for implementations.
Of course other calculi are possible for the same interpretation.
The plans for future work include:
\begin{itemize}
\item Investigations of further type constructions for a larger natural language coverage,
cf.\ the treatment of propositional attitudes in \cite{Villadsen01:LACL,Villadsen04:MATES}
which also replaces the classical logic with a paraconsistent logic.
\smallskip
\item Implementations using constraint solving technologies, cf.\ recent work on 
glue semantics \cite{Dalrymple99},
XDG (Extensible Dependency Grammar) \cite{Debusmann+04},
CHRG (Constraint Handling Rules Grammar) \cite{Christiansen02},
and categorial grammars \cite{deGroote01,Kuhlmann02,Moot99}.\!\!\!
\end{itemize}

\opt{x}{

\newpage

\section*{Appendix}

We also consider integrations of work concerning the ontology underlying natural language, to be specified in the lexicon, cf. as a starting point \cite{Dolling95}.

We are interested in a description of both syntax, semantics, and pragmatics of natural language.
As a brief illustration of the kind of semantic / pragmatic problems we have in mind we quote the famous ``fallacy of accent'' story:

\

\begin{small}
\noindent
~~\ldots~
Even the literal truth can be made use of, through manipulation of its placement, to deceive with accent.
Disgusted with his first mate who was repeatedly inebriated on duty, the captain of a ship noted in the ship's logbook,
almost every day, ``The mate was drunk today.''
The angry mate took his revenge.
Keeping the log himself on a day when the captain was ill, the mate recorded, ``The captain was sober today.''

\

\noindent
I. M. Copi \&\ C. Cohen (2002) \emph{Introduction to Logic} (11th ed.) Prentice Hall, p. 167.
\end{small}

\vfill

}

\providecommand{\crosscite}[1]{in \cite{#1}}


\opt{x}{

\vfill

\noindent
The numbers at the end of each bibliographical item above refer to the pages where the item is cited.

}

\end{document}